# Extension of Convolutional Neural Network with General Image Processing Kernels

Jay Hoon Jung, Yousun Shin, YoungMin Kwon
*Computer Science Department*
*Stony Brook University*
Stonybrook, New York
*Computer Science Department*
*The State University of New York Korea*
Incheon, South Korea
jay.jung@stonybrook.edu, yousun.shin@stonybrook.edu, youngmin.kwon@sunykorea.ac.kr

***Abstract*— We applied pre-defined kernels also known as filters or masks developed for image processing to convolution neural network. Instead of letting neural networks find its own kernels, we used 41 different general-purpose kernels of blurring, edge detecting, sharpening, discrete cosine transformation, etc. for the first layer of the convolution neural networks. This architecture, thus named as general filter convolutional neural network (GFNN), can reduce training time by 30% with a better accuracy compared to the regular convolutional neural network (CNN). GFNN also can be trained to achieve 90% accuracy with only 500 samples. Furthermore, even though these kernels are not specialized for the MNIST dataset, we achieved 99.56% accuracy without ensemble nor any other special algorithms.**

## I. INTRODUCTION

Feature extraction from images has long been regarded as an impossible task for any automata. Humans can instantly find a dog in an image without any efforts, but it is hard to set a rule or to provide mathematical tools for computers to do similar tasks. A breakthrough occurred in 2012, when Hinton's group showed that the convolutional neural networks can classify images from the ImageNet with a test error rate as small as 15.3%, which is far better than any other image processing programs at that time [1]. Since then deep neural networks have been used to extract features from the images without humans' helps.

Convolutional neural networks is a type of neural network. Instead of fully connected neural networks, CNN consists of neurons that are connected to local regions of an image. CNNs do not use predefined kernels, but it learns the kernels through training. According to Hinton, the network has learned 96 different kernels: frequency- and orientation-selective kernels as well as colored blobs. Interestingly, every run eventually ended up with similar types of kernels regardless of the initialization [1].

Although filter generation is a required processes for handling unknown images, it is also a bottleneck point of the entire learning process at the same time. Pre-trained kernels can help avoid this time consuming process [3], however they are not general-purpose filters, because they are specialized to input data features. Therefore, several algorithms such as Fast R-CNN [4], Faster R-CNN [5] and SSF-CNN [6] that are using pre-trained information show remarkable performances, but they are only applicable to trained image categories.

Digital image processing was adopted to send pictures cross the Atlantic Ocean in early 1920s. This technology could drastically reduce the intercontinental picture delivery time with a specialized system, but the image quality was not satisfactory. Researchers have devised diverse new ideas to improve the transmitted image quality, and image processing technology has rapidly grown since the advent of computers in 1960s. The computers offered enough storage space and computing power to implement researchers' ideas [7]. Many useful image processing filters have been developed. For example, Prewitt filter and Sobel filters were devised for edge detection. These filters have convolutional properties, and pixels are associated with adjacent pixels values. This concept is the same as the filters or kernels of CNN. The only difference is that the filter was applied to designed part for general image processing, whereas CNN filters are selected by the computer.

If two independent training of CNN resulted in a similar set of filters regardless of the initialization as Hilton's group showed, we can hypothesize that there might be a set of general filters for any CNN architecture. Meanwhile, we already have many filters developed in digital image processing: Sobel, Prewitt, and Robinson compass gradient operators, and blurring, sharpening and embossing filters, and discrete cosine transformation filters [7]. To exploit the capabilities of these filters, in this work, we substituted the first layer of CNN with these predefined filters. We expected that these general filters are useful to analyze images from MNIST so that they promote the performances of CNN.

## II. DATASET AND ARCHITECTURE

We designed GFNN as a general purpose method that can be used for any categories of images. Therefore, characters, symbols, natural creatures and artificial structures are possible to be inputs of GFNN. Since GFNN is an alternative method of randomly generated filters, it is independent from output types. In other words, GFNN is applicable to all neural networks if a network has convolutional layers.

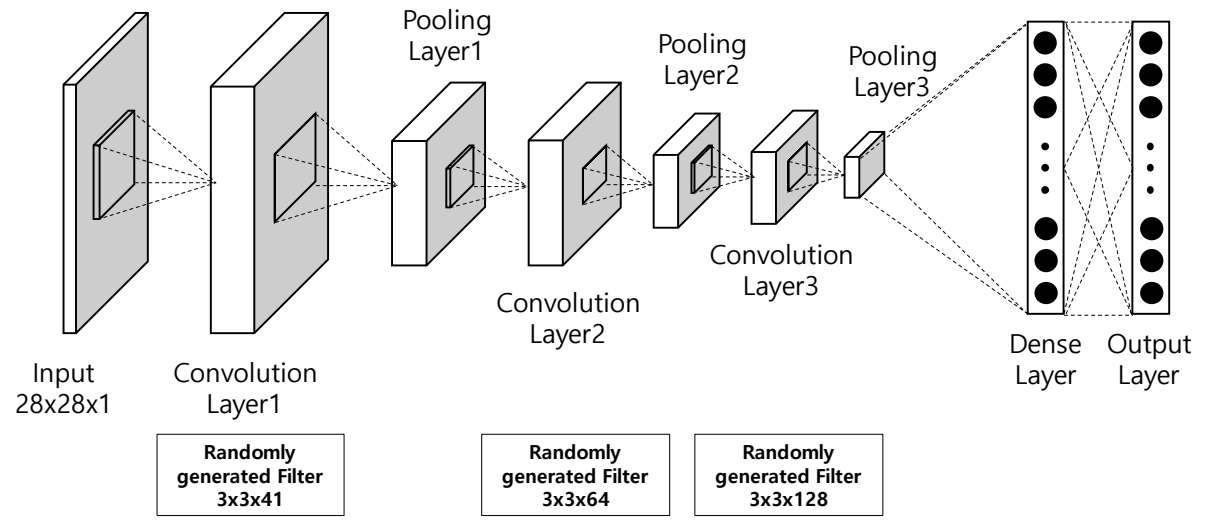

Fig. 1. Architecture of GFNN. The only difference with CNN is that we used 41 pre-defined 3x3 filters in the convolution layer 1.

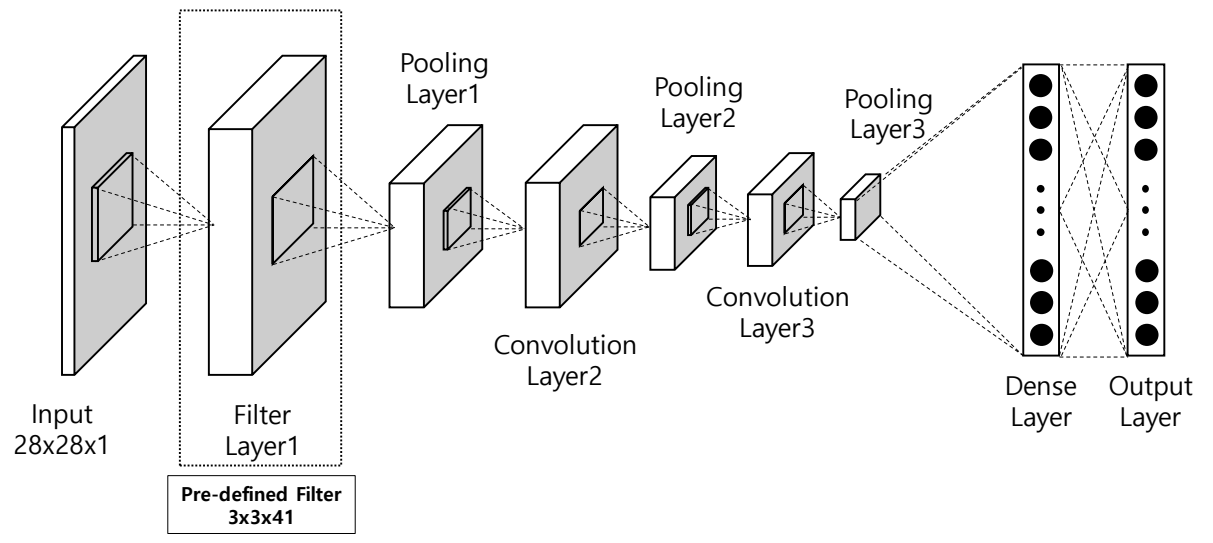

Fig. 2. Architecture of GFNN. The only difference with CNN is that we used 41 pre-defined 3x3 filters in the convolution layer 1.

## A. The Dataset: MNIST

The MNIST database is widely used for testing new image recognition algorithms in the machine learning community. It contains 60,000 handwritten images for training and 10,000 samples for testing. Since each image is a simple 28x28 one channel grayscale picture and is highly regularized [2], the MNIST is suitable for comparing the relative performance of the state-of-the-art algorithms. There are lots of well-known research results and training methods that are for increasing prediction accuracy with the MNIST. Among them, we chose the CNN using dropout as our baseline for the performance comparison.

## B. Convolutional Neural Network

Fig. 1 depicts an architecture of CNN. 28x28x1 images of handwritten numbers are feed into the network. Three convolution and max pooling layers are followed: 41 3x3 filters, 64 3x3 filters, and 128 3x3 filters are used for each convolution layer respectively with appropriate paddings to keep dimension of images. Convolution layer mimics the behavior of each neuron for visual stimulation and performs convolution operations. Max pooling reduces the amount of previous layers parameters by taking out non-critical parameters. We used 2x2 max pooling after convolution layer. The first and second dense layer has 2048x625 and 625x10 dimensions respectively. MNIST standards are followed when training the CNN except for the Fig. 6: 55000 training set and 5000 validation set. We changed the number of training samples for the last figure as indicated. We found that the learning rate of this CNN is optimal over different batch sizes and dropout rates when it is 0.001. Hence we used this value for rest of our training. All of the work in this paper is implemented using Tensorflow.

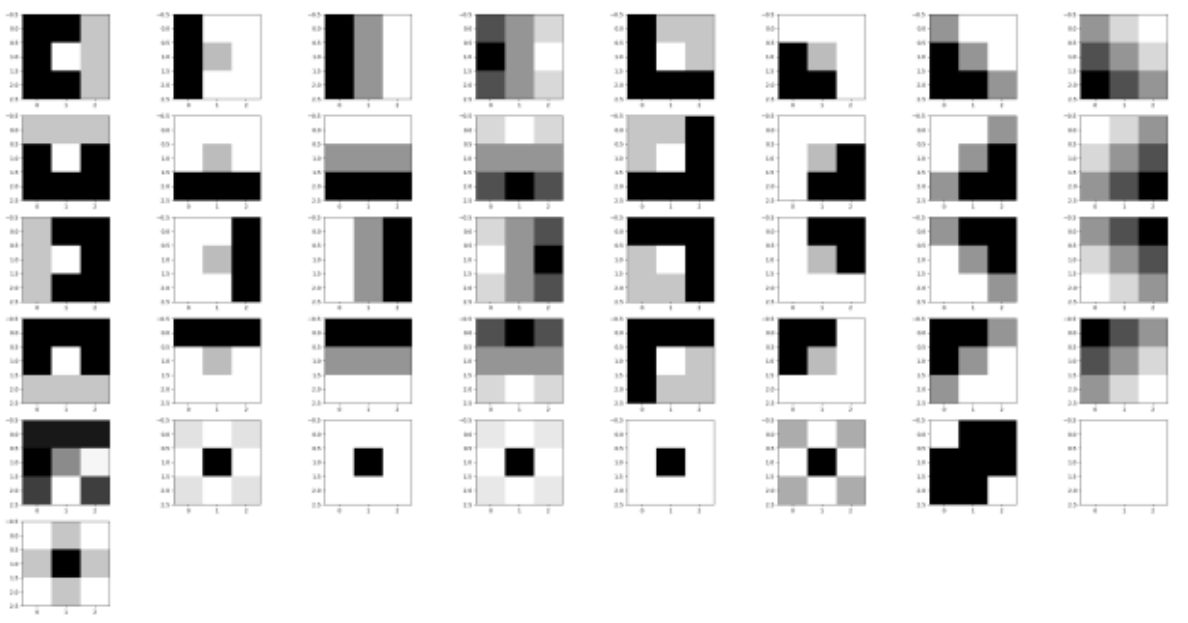

Fig. 3. A schematic view of 41 3x3 general kernels. 32 compass gradient operators , DCT filter, second order operators, sharpening, embossing, and blurring filters are drawn. Brighter color implies higher value.

## C. General Filter-Convolutional Neural Network

### 1) Architecture

Fig. 2 illustrates the architecture of GFNN. It has exactly the same structure of CNN in Fig.1 except for the first layer. For the first layer, named as filter layers, contains 41 pre-defined 3x3 kernels. All other parameters are consistent with the CNN as described in the previous section.

As the flow passes through each convolution layer, the targeting area of image becomes smaller. In other words, large images were processed at first, but it subsequently becomes smaller objects. Since MNIST data is composed of single numbers, we believed that processing large area has more significant effects than doing smaller objects. Thus, total 41 types of 3x3 matrices are applied to the first layer which dominates the whole network.

### 2) 41 3x3 General Filters

The filters consist of 3 sharpening filters, 1 embossing filter, 2 blurring filters, 32 edge detecting compass gradient operators, 2 second order operators, and 1 discrete cosine transformation operator (DCT) as shown in Table 1 and Fig. 3. We chose 32 edge detecting operators because we assumed that classification of MNIST dataset requires edge detection capabilities. We also assumed that sharpening, embossing, and blurring filters can help detecting edges when images do not have clear edges. We also assumed that DCT filters provide some useful information about the classification.

Fig. 3. shows a schematic view of these kernels. First 32 images are 32 compass gradient operators. The next are DCT, second order operators, sharpening, embossing, and blurring filters in that order.

TABLE I. GENRAL FILTER LIST

| Filter Name | | Number of filters |
|---|---|---|
| Sharpen | | 3 |
| Embossing | | 1 |
| Blur | | 2 |
| Compass Gradient Operators | Roberts | 8 |
| | Prewitt | 8 |
| | Sobel | 8 |
| | Frei-Chen | 8 |
| Second Order Operators | | 2 |
| Discrete Cosine Transformation Operator | | 1 |

## III. RESULT

In this section we compare the performances of GFNN and CNN. We compare them in terms of the training time, the accuracy of the result, and the sensitivity to the number of training samples.

### A. *Traning Time*

Fig. 4 shows that training time for 55000 samples of GFNN is 30% less than that of CNN on average. The main reason of the speed-up is the reduction of computing loads for training. By substituting convolution layers into filter layers, 41x3x3(=369) weight values are converted to 369 constant numbers. Although the number of eliminated weight variables is only 0.02% of the total number of weight values in the CNN, GFNN training time is significantly lower than that of CNN. It is presumably because the filter layer of GFNN do not have any weight values to update and each iteration of training can skip the first layer. In order words, GFNN performs multiplication between filter vectors and images only once during training. In the bar graph of Fig. 6, as the number of training samples increases, the training time gap between GFNN and CNN grows as well. Since the number of images in a batch do not change throughout the figure and the number of training samples increases along x-axis, the number of iteration increases too. In consequence, GFNN save time for the filter layer each iteration whereas CNN accumulates the computing time of the first layer.

### B. *Accuracy*

Even with the significant speed-ups, the accuracy of GFNN is higher compared to that of CNN as shown in Fig. 5 and 6. It seems that no correlation exists in between the number of images in a batch and the accuracy, but GFNN is more accurate regardless of the batch sizes as is indicated by Fig. 5. The average accuracy of GFNN is 99.42%, which is 0.2% higher than that of CNN. In addition, GFNN can reach 99.56% accuracy without any special methods such as ensembles or pre-training algorithms (data not shown here). In CNN, 99.5% accuracy is also achievable only after applying ensembles algorithms, but it adds much more time to complete the training.

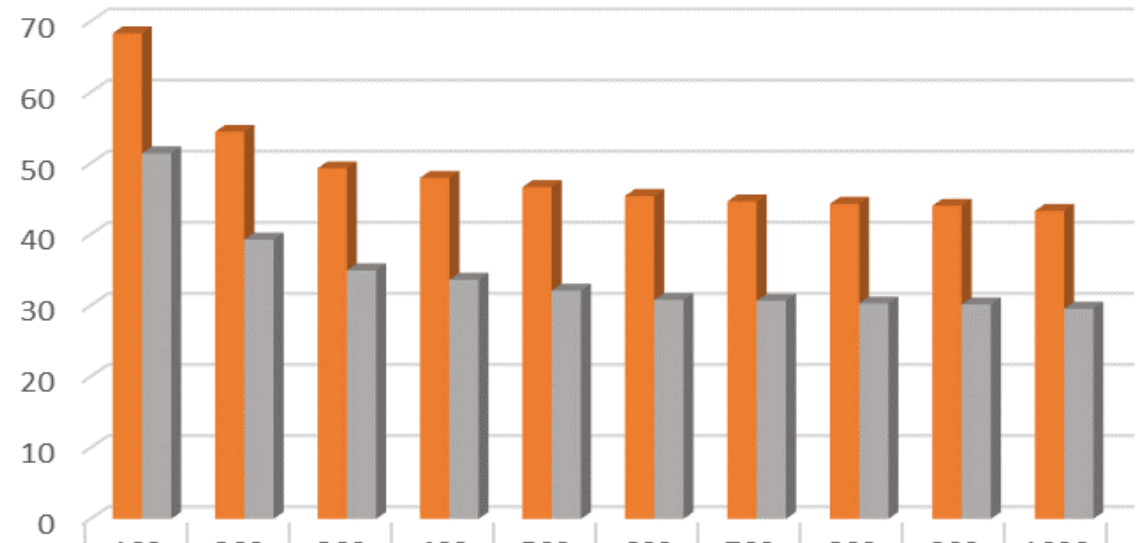

| | 100 | 200 | 300 | 400 | 500 | 600 | 700 | 800 | 900 | 1000 |
|---|---|---|---|---|---|---|---|---|---|---|
| CNN | 68.29 | 54.43 | 49.29 | 47.95 | 46.66 | 45.41 | 44.65 | 44.30 | 44.03 | 43.31 |
| GFNN | 51.41 | 39.28 | 34.96 | 33.64 | 32.10 | 30.80 | 30.68 | 30.28 | 30.17 | 29.56 |

Fig. 4. Training time comparison between CNN and GFNN. GFNN is 30% faster than CNN.

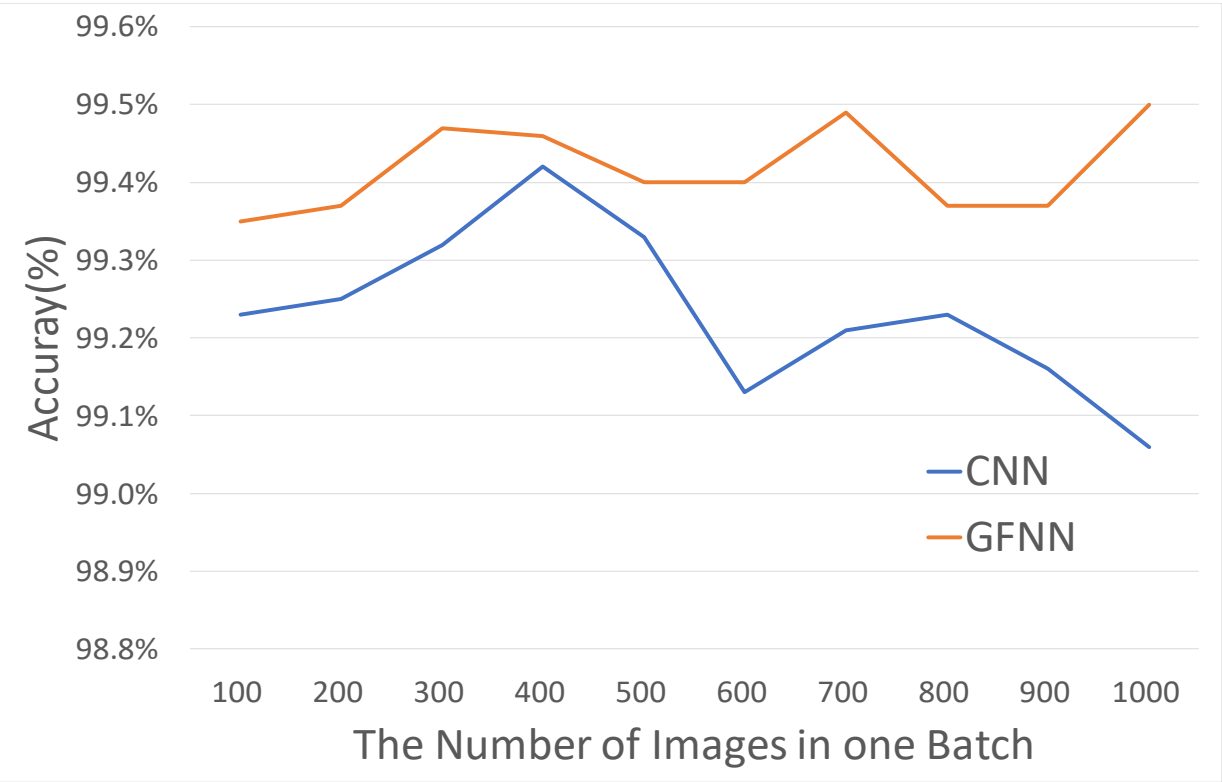

Fig. 5. Accuracy comparison between CNN and GFNN along with batch sizes. GFNN is more accurate than CNN.

### C. *Sensitivity to limited training set size*

GFNN reaches the accuracy of 90% even with only 500 training samples. CNN needs at least 2000 samples to get the 90% accuracy. Furthermore, the training time to get the same accuracy as GFNN takes twice longer than that of GFNN as shown in Fig. 6. It is worthwhile to mention that only 500 samples are used in the training, but GFNN can validate 5000 images with 90% accuracy, whereas CNN shows about 40% accuracy.

## IV. DISCUSSION

There are three advantages to use GFNN instead of CNN. Firstly, its training time is faster than regular CNN as is shown in the previous section. Secondly, GFNN needs fewer samples to train. Lastly, by investigating the list of filters, we can get some insights into how the neural network works. All these properties of GFNN address issues of neural networks: limited computing power, insufficient data, optimization of neural networks.

We only changed a single convolution layer to see the effects of the pre-defined filter set. As shown in Fig. 2, filters of GFNN are conventional filters used for general image processing. For instance, sharpness filters, the DCT filter and an embossing filter have been used since the early stage of computer vision technologies. In order words, filters of GFNN is independent to input images. It may limit the accuracy of the neural networks for certain datasets, but it will generically work for any arbitrary datasets.

GFNN can remove the bottleneck, the filter generation stage, so that it reduces the overall training time without compromising the accuracy. In other words, with the same computation power, we can deepen neural networks to have better performances.

The other advantage of GFNN is that it might be possible to build a neural network trainable with less samples than current CNNs. It is crucial to have large databases for neural networks. However, as GFNN suggested, with proper filters or data processing layers or units for neural networks, small training sample set might be enough. Researchers have developed many mathematical tools for humans to understand natures; why not let neural networks use those mathematical tools too instead of letting them find the tools again?

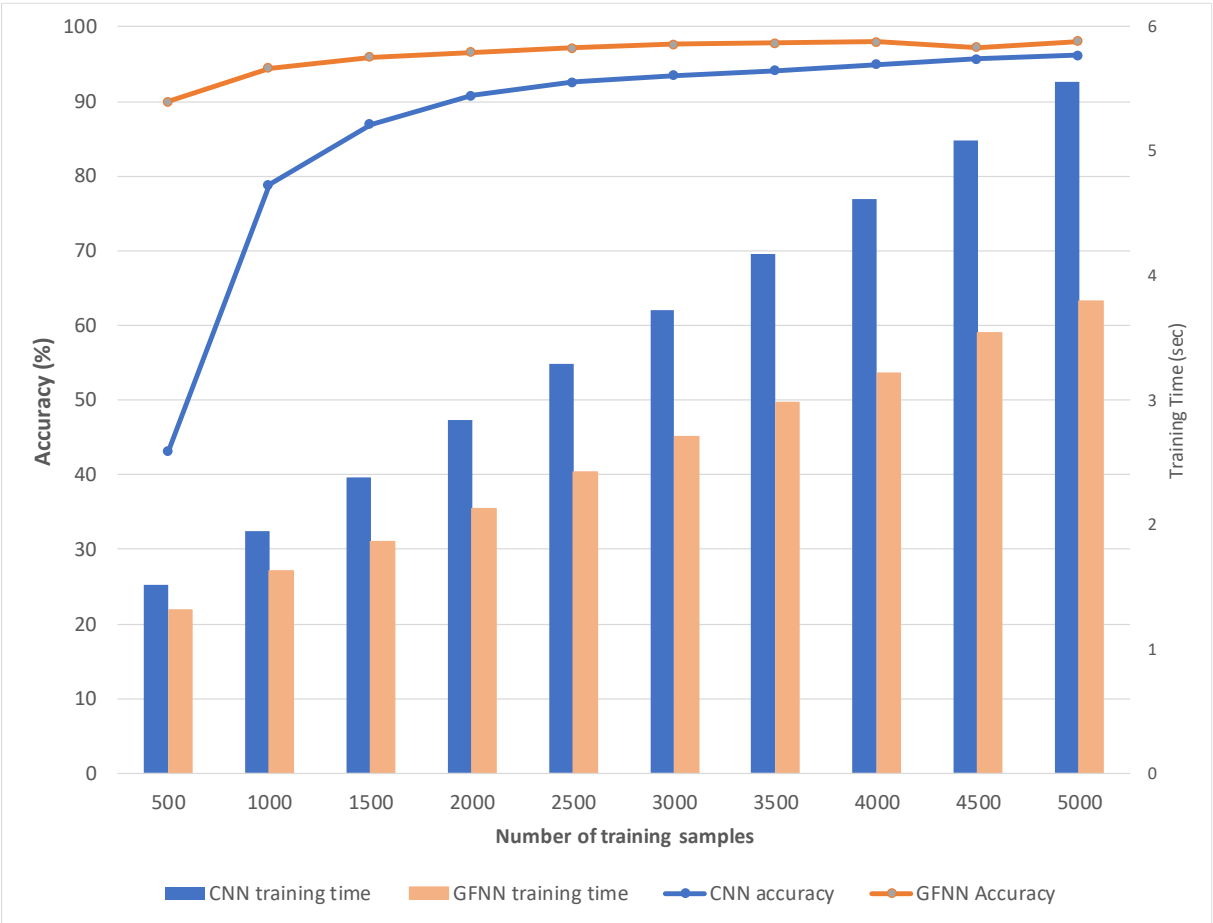

Fig. 6. Small sample size training and training time comparison between CNN and GFNN.

Lastly, it also may imply that if we want for neural networks to consider certain mathematical features, we can create such filter and add it to the filter set of GFNN. After the filter layer, the filter-processed image are directly fed into CNN. Neural networks will figure out the usefulness of the filter during the training; the weights related to the filter will have large values if the filter is making contributions. In addition, it might be possible to analyze how the neural network works by investigating those filter sets with large associated weights

## V. CONCLUSION

In this work, we substituted the first convolution layers of CNN into the filter layer, and named it as GFNN. Reducing repetitive computation, we achieved 30% faster training without degradation of accuracy. We also showed that GFNN can be trained only with 500 samples, and it produces 90% accuracy. Moreover, we discussed that we can use GFNN to extend CNN by applying different sets of filters.

## ACKNOWLEDGMENT

This research was partially supported by the MSIT (Ministry of Science and ICT), Korea, under the ICT Consilience Creative Program (IITP-2017-R0346-16-1007) supervised by the IITP (Institute for Information & Communications Technology Promotion) and partially by KEIT under the GATC program (10077300).